\documentclass[a4paper,11pt,final]{scrartcl}
\usepackage[utf8]{inputenc}
\usepackage{ifpdf}
\usepackage{latexsym}
\usepackage{exscale}
\usepackage{amsfonts}
\usepackage{eucal}
\usepackage[centertags]{amsmath}
\usepackage{amssymb}
\usepackage{amsbsy}
\usepackage{amstext}
\usepackage{amsthm,thmtools}
\usepackage{array,booktabs}%
\usepackage{mathtools}%
\usepackage{abstract}
\usepackage{xcolor}
\usepackage{framed}
\usepackage{tocbibind}
\usepackage{tocloft}
\usepackage{etoolbox}
\usepackage[ruled,vlined,algosection]{algorithm2e}
\usepackage{delarray}
\usepackage[
  automark,
  autooneside=false,
]{scrlayer-scrpage}
\clearpairofpagestyles
\ihead{\leftmark}
\ohead{\ifstr{\leftmark}{\rightbotmark}{}{\rightbotmark}}
\usepackage{lmodern}
\usepackage[toc,page]{appendix}
\usepackage{textcomp}
\usepackage{mflogo}
\usepackage{here}
\usepackage{listings}
\usepackage[round]{natbib}
\usepackage{here}
\usepackage{longtable}
\usepackage{threeparttablex}
\usepackage{multirow}
\usepackage{multicol}
\usepackage{caption}
\usepackage{lscape}
\usepackage{rotating,capt-of}
\usepackage[margin=15mm]{geometry}
\usepackage{pgf}
\usepackage{tikz}
\usetikzlibrary{arrows,automata,shapes,matrix}
\usepackage{verbatim}
\newcounter{treeline}

\ifvtexpdf\pdftrue\fi
\ifpdf
\usepackage{graphicx}
\usepackage[pdftex]{thumbpdf}
\usepackage{nameref}
\usepackage[pdftex,colorlinks]{hyperref}
\hypersetup{
           breaklinks=true,
           pdfpagemode=UseOutlines,
           pdfstartview=FitH,
           colorlinks=true,
           linkcolor=blue,
           citecolor=blue,
           bookmarksopen=true,
           pdftitle={Deduction Theorem: The Problematic Nature of Common Practice in Game Theory},
           pdfauthor={Holger Ingmar Meinhardt},
           pdfsubject={TU Games, Cooperative Game Theory, Industrial Cooperation},
           pdfkeywords={Propositional Logic, Deduction Theorem, Herbrand Theorem, Proof by Contradiction, TU Games, Cooperative Oligopoly Games, Partition Function Approach, Nash Equilibrium, Aggregation across Firms},            
}

\else
\usepackage{epsfig}
\usepackage[ps2pdf]{thumbpdf}
\usepackage{nameref}
\usepackage[ps2pdf]{hyperref}

\hypersetup{
           breaklinks=true,
           pdfpagemode=UseOutlines,
           colorlinks=true,
           linkcolor=blue,
           citecolor=blue,
           bookmarksopen=true,
           pdftitle={Deduction Theorem: The Problematic Nature of Common Practice in Game Theory},
           pdfauthor={Holger Ingmar Meinhardt},
           pdfsubject={TU Games, Cooperative Game Theory, Industrial Cooperation},
           pdfkeywords={Propositional Logic, Deduction Theorem, Herbrand Theorem, Proof by Contradiction, TU Games, Cooperative Oligopoly Games, Partition Function Approach, Nash Equilibrium, Aggregation across Firms}, 
}
\fi

\addtocounter{MaxMatrixCols}{20}

\DeclareCaptionLabelSeparator{spacing}{:\quad}
\newlength{\tabularextravskip}
\newcolumntype{M}[1]{>{\centering\arraybackslash$}m{#1}<{$}}

\makeatletter
\renewcommand{\section}{\scr@startsection
  {section}%
  {1}%
  {0em}%
  {-\baselineskip}%
  {0.5\baselineskip}%
  {\centering\normalfont\Large\scshape\mdseries}}%
\makeatother

\makeatletter
\renewcommand{\subsection}{\scr@startsection
  {subsection}%
  {2}%
  {0em}%
  {-\baselineskip}%
  {0.5\baselineskip}%
  {\normalfont\large\scshape\mdseries}}%
\makeatother

\makeatletter
\renewcommand*\env@matrix[1][c]{\hskip -\arraycolsep
  \let\@ifnextchar\new@ifnextchar
  \array{*\c@MaxMatrixCols #1}}
\makeatother

\makeatletter

\newenvironment{theopargself*}
    {\def\@spopargbegintheorem##1##2##3##4##5{\trivlist
         \item[\hskip\labelsep{##4##1\ ##2}]{\hspace*{-\labelsep}##4##3\@thmcounterend}##5}
     \def\@Opargbegintheorem##1##2##3##4{##4\trivlist
         \item[\hskip\labelsep{##3##1}]{\hspace*{-\labelsep}##3##2\@thmcounterend}}}{}
\def \@floatboxreset {%
        \reset@font
        \small
        \@setnobreak
        \@setminipage
}
\def\figure{\@float{figure}}
\@namedef{figure*}{\@dblfloat{figure}}
\def\table{\@float{table}}
\@namedef{table*}{\@dblfloat{table}}
\def\fps@figure{htbp}
\def\fps@table{htbp}

\makeatother

\makeatletter
\@addtoreset{equation}{section}
\makeatother
\makeatletter
\@addtoreset{table}{section}
\makeatother

\theoremstyle{plain}

\newtheorem{statement}{Statement}[section]

\newtheoremstyle{break}
  {9pt}
  {9pt}
  {\itshape}
  {}
  {\bfseries}
  {.}
  {\newline}
  {}

\newtheoremstyle{break1}
  {9pt}
  {9pt}
  {\rmfamily}
  {}
  {\scshape}
  {.}
  {\newline}
  {}

\theoremstyle{break}

\newtheoremstyle{note}
  {3pt}
  {3pt}
  {}
  {}
  {\itshape}
  {:}
  {.5em}
  {\newline}  
  {}

\theoremstyle{note}

\theoremstyle{definition}

\theoremstyle{break1}

\begin{document}

\bibliographystyle{plainnat} 
\pdfbookmark[0]{Deduction Theorem: The Problematic Nature of Common Practice in Game Theory }{tit}

\title{{Deduction Theorem: The Problematic Nature of Common Practice in Game Theory}  
}
\author{{\bfseries Holger I. MEINHARDT}~\thanks{The author acknowledges support by the state of Baden-Württemberg through bwHPC. Of course, the usual disclaimer applies.}
~\thanks{Holger I. Meinhardt, Institute of Operations Research, Karlsruhe Institute of Technology (KIT), Englerstr. 11, Building: 11.40, D-76128 Karlsruhe. E-mail: \href{mailto:Holger.Meinhardt@wiwi.uni-karlsruhe.de}{Holger.Meinhardt@wiwi.uni-karlsruhe.de}} 
}
\maketitle

\begin{abstract}
We consider the Deduction Theorem used in the literature of game theory to run a purported proof by contradiction. In the context of game theory, it is stated that if we have a proof of $\phi \vdash \varphi$, then we also have a proof of $\phi \Rightarrow \varphi$. Hence, the proof of $\phi \Rightarrow \varphi$ is deduced from a previously known statement. However, we argue that one has to manage to establish that a proof exists for the clauses $\phi$ and $\varphi$, i.e., they are known true statements in order to show that $\phi \vdash \varphi$ is provable, and that therefore $\phi \Rightarrow \varphi$ is provable as well. Thus, we are not allowed to assume that the clause $\phi$ or $\varphi$ is a true statement. This leads immediately to a wrong conclusion. Apart from this, we stress to other facts why the Deduction Theorem is not applicable to run a proof by contradiction. Finally, we present an example from industrial cooperation where the Deduction Theorem is not correctly applied with the consequence that the obtained result contradicts the well-known aggregation issue.\\ 

\noindent {\bfseries MS Classifications 2010}: 03B05, 91A12, 91B24  
\end{abstract}

\thispagestyle{empty}

\pagestyle{scrheadings}  \ihead{\empty} \chead{Deduction Theorem: The Problematic Nature of Common Practice} \ohead{\empty}

\section{Introduction}
\label{sec:intro}

We review a common practice in the literature of game theory of applying the Deduction Theorem (Herbrand Theorem, 1930) on a purported proof by contradiction. This theorem is a meta-theorem of mathematical logic, which states that if a known proposition/clause $\varphi$ within a calculus system is deducible (provable) by some known clauses $\phi_{1},\ldots,\phi_{n}$ from the calculus system -- written more formally as $\phi_{1},\ldots,\phi_{n} \vdash \varphi$, then the derived term within the calculus system is also provable. Notice that the calculus system describes a logical consistent and complete rule (a calculus) of how we can get new logical terms (clauses) from already known true statements. It is the closed logical framework/system from which we obtain new results. Or to state it differently, a calculus system describes the axiomatic theory of propositional logic. For a more thorough discussion, we refer the reader to~\citet{ebb:07}.    

In the literature of game theory, authors apply a reduced form of the Deduction Theorem, in particular, while trying to run an indirect proof. In this context, it is stated that if we have a proof of $\phi \vdash \varphi$, then we also have a proof of $\phi \Rightarrow \varphi$. Thus, if we have managed to prove the clauses $\phi$ and $\varphi$, they are known true statements of a calculus system from which the provability of $\phi \vdash \varphi$ can be obtained.\footnote{By abuse of language, we state sometimes in the course that a clause $\phi$ exists rather to say that a proof of the clause $\phi$ exists in a calculus system.} This means that if we have successfully reasoned on the provability of $\phi \vdash \varphi$, then the derived judgment that $\phi \Rightarrow \varphi$ is provable can be deduced. Unfortunately, this is not the common practice observed in game theory.\footnote{Personally, we have the impression that the great majority of game theorist is not really aware that they base their argumentation on the Deduction Theorem.} In contrast, authors impose existence assumptions on the provability of $\phi$ and $\varphi$ to reason that $\phi \vdash \varphi$ is provable, and that this proves $\phi \Rightarrow \varphi$. By the foregoing reasoning it should be clear that we are only allowed to reason with known true statements, that is, we are not allowed to assume that a proof for $\phi$ or $\varphi$ exists. Doing so, leads immediately to a wrong conclusion.            

The remainder of the treatise is organized as follows: Section~\ref{sec:logstat} provides a short refresher of some essential aspects of propositional logic. Having discussed this, we are going to discuss in more detail within Section~\ref{subsec:deduc_thm} the incorrect application of the Deduction Theorem while trying to run a proof by contradiction. Whereas Section~\ref{subsec:undefga} is dedicated to an example from the literature, where an author derives results while applying the Deduction Theorem incorrectly, that is, imposing an existence assumption that contradicts the well-known issue of bequeathing essential properties of a function from an individual to an aggregate level. We close our considerations with some final remarks in Section~\ref{sec:rem}.      
 
\section{Propositional Logic: A Refresher}
\label{sec:logstat}

For presenting a reminder of propositional logic, we introduce two truth tables. A logical statement/proposition is formed by the symbols $A$ or $B$, which means that a statement $A$ is true or false. However, the inversion is formed by the negation of a proposition by using the logical term ``not'' denoted by $\neg$. If $A$ is a proposition, then $\neg A$ is the negation of $A$ verbalized as ``not $A$'' or ``$A$ is false''. The effect of negation, conjunction, disjunction, and implication on the truth values of logical statements is summarized by a so-called truth table. In this table, the capital letter {\bfseries T} indicates a true proposition and {\bfseries F} indicates that it is false. 

\begin{center}
\begin{tabular}{c c c c c c c c c c c}
\hline
$A$ & $B$ & $\neg B$ & $A \Rightarrow B$ & $ \neg (A \Rightarrow B)$ & $A \Leftarrow B$ & $A \Leftrightarrow B$ & $A \lor \neg B$ & $A \land B $ &  $A \lor B $\\
\hline
F & F & T & T & F & T & T & T & F & F \\
F & T & F & T & F & F & F & F & F & T \\
T & F & T & F & T & T & F & T & F & T \\
T & T & F & T & F & T & T & T & T & T \\ 
\end{tabular}

\begin{tabular}{c c c c |c c ||c c| c c}
\hline
$A$ & $B$ &  $\neg A$ & $\neg B$ & $\neg A \Rightarrow \neg B$ & $A \lor \neg B$ & $\neg A \Leftarrow \neg B$ & $\neg A \lor B$  & $A \land \neg B$ & $\neg A \Leftrightarrow \neg B$ \\
\hline
F & F & T & T & T & T & T & T & F & T   \\
F & T & T & F & F & F & T & T & F & F   \\
T & F & F & T & T & T & F & F & T & F   \\
T & T & F & F & T & T & T & T & F & T   \\\hline
\end{tabular}
\end{center}
Two statements are indicated as logically equivalent through the symbol $\equiv$. For instance, by the truth table we realize that the two statements $\neg A \Leftarrow \neg B$ and $\neg A \lor B$ are logically equivalent, which is formally expressed by $(\neg A \Leftarrow \neg B) \equiv (\neg A \lor B)$. A falsum $\bot$ is, for instance, the conjunction $A \land \neg A$ whereas a tautology $\top$ can be expressed, for instance, by the disjunction $\neg A \lor A \equiv \neg(A \land \neg A)$. Moreover, a proposition or premise $A$ might satisfy a falsum or a tautology or an arbitrary property $B$, which is expressed by $(A \vdash \bot)$ or $(A \vdash \top)$ or $(A \vdash B)$ respectively. This should not be confounded with an implication of the form $(A \Rightarrow \bot)$ or $(A \Rightarrow \top)$ or $(A \Rightarrow B)$ respectively.

Finally, to capture a material implication or adjunctive logical statements by a concise notation, we also refer to $\phi$ or to $\varphi$, and to a system of terms by $\Phi$. By the latter, we define a system of terms $\Phi$ as consistent if, and only if, there exists no term $\phi$ s.t. $\Phi \vdash \phi$ and $\Phi \vdash \neg \phi$ hold simultaneously. In contrast, a system of terms $\Phi$ is said to be inconsistent if, and only if, there exists a term $\phi$ s.t. $\Phi \vdash \phi$ and $\Phi \vdash \neg \phi$ hold simultaneously. Moreover, a system of terms $\Phi$ is inconsistent if, and only if, for all $\phi$ it holds $\Phi \vdash \phi$ ({\bfseries Principle of Explosion}). This statement is  logically equivalent to the statement that a system of terms $\Phi$ is consistent if, and only if, there exists a $\phi$ that cannot be deduded by $\Phi$ (cf.~\citet[Chap.~4.]{ebb:07}). A material implication is a rule of replacement that allows to replace a conditional proposition by a disjunction. For instance, the conditional statement $A$ implies $B$ can be replaced by the disjunction $\neg A \lor B$, which is logically equivalent to the former proposition (see the truth table).

These information are enough to discuss some formal representations of an indirect proof, which is also called a proof by contradiction. This proof technique enables us to prove a proposition $A$ by establishing that the negation $\neg A$ implies something that we know to be false. Hence, the negation $\neg A$ contradicts a known true statement.  
 
To show that a proposition $A$ and its negation $\neg A$ cannot both be true, some uses the representation of a conditional statement like $\neg A \Rightarrow A$ to conclude that $\neg A$ and $A$ cannot both be true, and that a contradiction occurs with the consequence that the initial assumption must be wrong and $A$ must be true. But this is fallacy, which confounds $\neg A \Rightarrow A$ with $\neg A \vdash  A$. First of all, $\neg A \Rightarrow A$ simply says that if $\neg A$ is false then $A$ must be true, and if $\neg A$ is true, then $A$ has to be false. On this statement is nothing contradictory. Moreover, we do not have $\neg A \land A$ as we can observe by the following chain of equivalences
\begin{equation*}
  (\neg A \Rightarrow A) \equiv (\neg (\neg A) \lor A) \equiv (A \lor A) \equiv A. 
\end{equation*}
Hence, $\neg A \Rightarrow A$ is equivalent to $A$, but not to $\neg A \land A$. In contrast, a representation of an indirect proof by $\neg A \vdash A$ indicates that the initial assumption made by $\neg A$ is false, since $\neg A$ satisfies $A$. Here one deduces that $\neg A$ cannot hold, the initial assumption is wrong and $A$ holds. But this is not an implication, this is a deduction. This reveals that deduction and implication are not synonymous expressions.

Alternatively, to involve that a proposition $\phi$ ``satisfies'' a falsum $\bot $ in order to conclude that in fact $\neg \phi$ holds and a contradiction is derived, we write $(\phi \vdash \bot) \Leftrightarrow \neg \phi$. This constitutes a formal expression of an indirect proof. Notice that this has per se nothing to do with an application of the forthcoming Deduction Theorem, hence it is allowed to assume that $\phi$ is a true statement. Be aware that imposing the assumption of the existence of a proof for $\phi$ is not anymore allowed if we apply the Deduction Theorem to reason that $(\phi \vdash \bot)$ is provable to finally conclude that $(\phi \Rightarrow \bot)$ is provable as well. Moreover, it should be evident that this is not the same as $(\phi \Rightarrow \bot) \Leftrightarrow \neg \phi$. Since in the former case we have that a proposition $\phi$ ``satisfies'' $\bot$ whereas in the latter case a proposition $\phi$ ``implies'' $\bot$.      
 
In the context of a calculus system, we have to take into account that if a proposition is provable (deducible), then it is a correct (true) logical statement, and if it is a true logical statement, then it is provable. Saying that in a calculus system all logical true statements are provable, and vice versa.

\section{The Deduction Theorem}
\label{subsec:deduc_thm}
The Deduction Theorem (Herbrand Theorem, 1930) is roughly stating that each judgment of a system of terms $\Phi$ must be logically independent and consistent, i.e.~the judgment must be based on a calculus system, in order to conclude that the derived judgment is deducible (cf.~\citet[Chap.~11.]{ebb:07}). This means, for instance, that if we have a proof of $\phi \vdash \varphi$, then we also have a proof of $\phi \Rightarrow \varphi $. Notice that if the clause $\phi \vdash \varphi $ is provable, then $\phi$ and $\varphi$ must be consistent and logically independent clauses of a system of terms $\Phi$. Here, the term consistent can be interpreted as indicating provable as well, since we know that $\Phi \vdash \phi$ if, and only if, $\Phi \cup \{\neg \phi\}$ is not consistent, or to put it differently, they are known true statements, saying that a proof exists for the clauses $\phi$ and $\varphi$. Apparently, if $\phi$ and $\varphi$ are clauses in a calculus system then they are consistent and logically independent clauses from which the provability of $\phi \vdash \varphi$ on the calculus system can be derived. Hence, $\phi$ and $\varphi$ can only be used on a system of terms, if we have managed to prove them. Thus, to prove $\phi \vdash \varphi $, we are only allowed to reason with known true statements, that is, we are not allowed to assume the existence of a prove for $\phi$, for instance; otherwise this leads immediately to a wrong conclusion (see also the forthcoming discussion of Section~\ref{subsec:undefga}).      

This is caused by the fact that a game theoretical analysis introduces a system of terms whose completeness and consistency must be established, i.e.,~that it is a calculus system. Thus, introducing a system of terms does not assure the existence of a calculus system, it is just a prototype or candidate of a calculus system. The problem of working with a prototype of a calculus system can be explained on the basis that it is not allowed to make an assumption that the clauses $\phi$ and $\varphi$ are true statements. Then, apparently, not all logical true statements are provable violating the consistency of the prototype of the calculus, and the existence of calculus must be denied. Moreover, in such a case the logical independence of $\phi$ and $\varphi$ cannot be guaranteed implying that a proof by contradiction based on $\phi \vdash \varphi $ is useless whenever they are logically dependent. As a result the system of terms is inconsistent. Therefore, from an inconsistent system incorrect conclusions were derived.           

By this consideration, it is only possible to get a derived deduction from a system of clauses with known proofs. Saying that from a consistent and closed system, i.e.,~a calculus, containing $\phi$ and $\varphi$, a deduction $\phi \vdash \varphi $ can be obtained, from which we can infer that the clause $\phi \Rightarrow \varphi $ is provable. However, it is common practice in the literature of game theory that authors try to deduce $\phi \vdash \varphi $ while assuming that $\phi$ is true with one possible false statement $\varphi$, in order to conclude that $\phi \Rightarrow \varphi $ is provable. It should be evident that this contradicts the foregoing described procedure to correctly apply the Deduction Theorem, and therefore violates good practice of logical reasoning.           
 
\subsection{Incorrect Application of the Deduction Theorem}
\label{subsec:incor_deduc}
 
By the foregoing discussion we have argued that the Deduction Theorem was not correctly applied in the literature of game theory. In particular, as we shall see in a short while, this practice finds wide application to run a purported proof by contradiction. In the course, we shall now take the opportunity to refute by a more thorough reasoning the arguments which are presented in favor of its application in the game theory literature. 

There is a growing literature of game theory (cf.~\citet{mei:15,mei:16,mei:16b,mei:17,mei:18a,mei:18b}), where one can find incorrect applications of the indirect proof w.r.t.~a logical sentence like 
\begin{equation}
  \label{eq:start}
A  \Rightarrow B,
\end{equation}
that cannot be managed to prove it directly. It is argued that the sentence~\eqref{eq:start} is, nevertheless, provable by assuming that this is a wrong judgment while establishing that the equivalent implication given by
\begin{equation}
  \label{eq:logindp0}
(A \land \neg B) \Rightarrow (A \land \neg A),
\end{equation}
is a valid logical sentence. Hence, these authors claimed that such a logical sentence is deducible, and that it is therefore a correct logical judgment. A fortiori, we shall establish that this is not a provable statement. That means, there exists no valid proof for the implication~\eqref{eq:logindp0}.   

Recall, in order to apply the Deduction Theorem we have to take into account that if a proposition is provable (deducible), then it is a correct (true) logical statement, and if it is a true logical statement, then it is provable. In the course of this discussion, we shall now investigate the issue for~\eqref{eq:logindp0} in more detail.  

In the following, we first demonstrate that this implication is not a true logical statement in order to apply the Deduction Theorem. To see this, we redefine the implication by setting 
\begin{equation}
  \label{eq:def}
  (A \land \neg B) := \neg(A \Rightarrow B) \qquad (A \land \neg A) :=\neg(A \Rightarrow A).
\end{equation}
Then, the implication of~\eqref{eq:logindp0} can be rewritten as 
\begin{equation}
  \label{eq:rw}
\neg(A \Rightarrow B) \Rightarrow \neg(A \Rightarrow A).
\end{equation}
By the assumption made in the literature that $A  \Rightarrow B$ is a false judgment, which is the case whenever $A$ is true i.e., $T$, and $B$ is false $(F)$, we get
\begin{equation}
  \label{eq:rw2}
  \begin{split}
\neg(A \Rightarrow B) & \Rightarrow \neg(A \Rightarrow A) \\
\neg(T \Rightarrow F) & \Rightarrow \neg(A \Rightarrow A) \\
\neg(F) & \Rightarrow \neg(T \Rightarrow T) \\
 T & \Rightarrow \neg(T) \\
 T & \Rightarrow F \\
 & F.
 \end{split}
\end{equation}
Thus, we observe by these arguments that the implication~\eqref{eq:rw} is not a valid logical statement. From the point of view that this is not a true logical statement, it is also not provable.

At the final step let us check if the expression $\neg(A \Rightarrow B)$ satisfies $\neg(A \Rightarrow A)$ i.e., $\neg(A \Rightarrow B) \vdash \neg(A \Rightarrow A)$ is a deduction, for which authors give no rigorous logical argument, instead of that they refer to fatal logical arguments to make their point that~\eqref{eq:logindp0} is provable while applying for $(\phi \Rightarrow \bot) \Leftrightarrow \neg \phi$ incorrect arguments. 

To answer this issue, we note first that it is well known that $A \Rightarrow B$ is a deducible expression, but then the negation of $A \Rightarrow B$, the expression $\neg(A \Rightarrow B)$ cannot be deducible. Similar, we know that $A \Rightarrow A$ is a deduction, but then the negation of $A \Rightarrow A$ i.e., $\neg(A \Rightarrow A)$, is not deducible as well. From these findings, it is obvious that there exists no deduction $\neg(A \Rightarrow B)  \vdash \neg(A \Rightarrow A)$ at all, and therefore no proof. Otherwise, one would say that a non deducible expression can be deduced by something which is not deducible. This is obviously absurd. Moreover, what does it mean if we could deduce something from an inconsistency? Now, this would simply mean that from something inconsistent one can deduce everything, therefore we would have $\neg (A \Rightarrow B) \vdash C$, for every $C$, including $C := \neg (A \Rightarrow A)$.\footnote{This is the {\bfseries Principle of Explosion} also subsumed under ex falso sequitur quodlibet,~i.e., from falsehood anything follows (cf.~Section~\ref{sec:logstat}). Is the antecedence (premise) inconsistent, it cannot be true, then drawing a conclusion is meaningless, and anything can be concluded. Hence, this says that an inconsistent theory -- for instance, an inconsistent game model -- is useless.} Of course, this also is an absurd judgment. Moreover, it is mandatory to establish that the candidate of a calculus system is logically independent, which means in that case, one has to show that the prerequisites $A$ and $B$ are logically independent. From these findings, we may conclude that there exists no valid deduction $\neg(A \Rightarrow B)  \vdash \neg(A \Rightarrow A)$, and the expression is therefore not provable. Since this is not a deduction, it is also not a true logical statement.           
 
Summarizing our results: First, we have established that the Deduction Theorem is not applicable. Second, we may even conclude that there exists no proof for $(A \land \neg B) \Rightarrow (A \land \neg A)$. Hence, there is no deduction $\neg(A \Rightarrow B) \vdash \neg(A \Rightarrow A)$ to conclude that $\neg(A \Rightarrow B)  \Rightarrow \neg(A \Rightarrow A)$ is a correct logical statement. 
  
\section{Aggregation across Firms: A Fallacy} 
\label{subsec:undefga}

In the course of this Section, we shall develop by means of an example that wrong conclusions will be made whenever the Deduction Theorem is not correctly applied. Conclusions that are inconsistent to known true statements from the literature as in the forthcoming example. 

For this purpose, we have selected the work of~\citet{lard:12}, who claimed by Proposition 3.1 that each characteristic function derived from a partition function approach is well defined. We shall establish that this result contradicts aggregate behavior that is well known from the theory of consumers (see~\citet{Russell:83,var:92,Jerison:93,Russell:93,Russell:96,Russell:98}), which is, however, also a relevant issue for the theory of industrial cooperation, since one has to aggregate across firms. Even in the latter case the aggregate profit function must satisfy the crucial properties of quasi-concavity and continuity. As we shall learn in a while, the second property cannot be guaranteed on an aggregated level while introducing regular conditions (cf.~\citet{nash:51,VDa:96,msz:13}) like compactness and convexity of the strategy set; and continuity as well as quasi-concavity of player's profit function. To be more precise, continuity of the individual profit function of a normal form game is a necessary, but not a sufficient condition of continuity of a profit function of a derived aggregate normal form game. By the mentioned literature from consumer theory, we know that consumers utility functions must be of Gorman form\index{Gorman form} to get an aggregate demand function that inherits their properties. This type of utility function is quasi-linear, or to put it differently, it is linear in the numeraire good. A property that cannot be guaranteed for an aggregate profit function of an industry, and that cannot assure either that aggregate best reply functions are continuous. However, by pointing to this reference from consumer theory, we notice that a particular form of individual profit functions is needed to bequeath their properties to the aggregate profit function of an cartel, for instance.                   
 
In order to work out the conflict to~\citet[Proposition 3.1]{lard:12}, we focus on the $\gamma$-characteristic function -- that is, outsiders' behavior is characterized by the $\gamma$-belief to specify the partition function from which finally the characteristic function is derived (see~\citet{hartku:83}, and for an application in industrial cooperation~\citet{mei:17f}) -- to establish that this function may not be well-defined, and that therefore a Nash equilibrium (partial agreement equilibrium) for a particular coalition structure based on oligopoly situation without transferable technologies cannot be expected to exist while aggregating across firms. The conflict is caused by an incorrect application of the proof by contradiction that fades out the aggregation issue with the fatal consequence that the author concludes that $\gamma$-characteristic function is well-defined, and that the Nash-equilibrium always exists for normal form games w.r.t.~a partition of the player set that is based on the $\gamma$-belief. Which is a fallacy, since the intrinsic aggregation issue clearly excluded this possibility. Thus, there may exist no Nash equilibrium for that specific coalition structure, which contradicts the assertion of Proposition 3.1. An effect which is for Cournot oligopoly games only visible due to combinatorial reasons from five firms onward. This issue shall be investigated in greater detail by the forthcoming counter-example.     

\subsection{Investigation of the Arguments}
\label{subsec:invarg}
During this discussion, we quote his fatal logical arguments while running a purported indirect proof that leads the author to wrong conclusion. Notice, we only summarize the main arguments of the author without going into the details, and without discussing the notation as well as the definitions. To start with, we recall~\citet[Proposition 3.1]{lard:12} and quoting his proposition as a simple statement to mark the incorrectness.  

\begin{statement}
  \label{lard12:prop3d1}
  Let $\langle N, (X_{k},\pi_{k})_{k \in N} \rangle $ be a normal form game. Then for any coalition structure $\mathcal{P} \in \Pi(N)$, there exists an equilibrium under $\mathcal{P}$. 
\end{statement}

Some arguments, we are going to present here, have been already developed within the work of~\citet{mei:15,mei:16,mei:16b,mei:17}. Nevertheless, we present some refinements and extensions of our argumentation that go beyond those we have in particular been applied in~\citet[Section 4]{mei:15}. For a quick reminder of propositional logic, the inclined reader ought consult Section~\ref{sec:logstat}.   

In the article of~\citet{lard:12}, the author claims to provide for the class of oligopoly TU games an existence result of the $\gamma$-core and a single-valued allocation rule inside of the $\gamma$-core that is called by the author Nash Pro rata-value. Moreover, he asserts to present an axiomatic characterization of the NP-value. Nevertheless, this article is strongly flawed due to the fact that the author confuses and mixes up non-equivalent fundamental statements from propositional logic while applying false indirect arguments.

In particular, the author deduces wrong conclusions from logical statements derived from an indirect proof which relies on a material implication. In this context, he neither recognizes the logical relationship $(A \land \neg B \Rightarrow A \land \neg A) \equiv (A \Rightarrow B)$ nor $(\neg A \land B \Rightarrow \neg A \land A) \equiv (B \Rightarrow A)$. Moreover, he does not notice that the left hand side of these equivalence relations is in each case an inappropriate application of the Deduction Theorem (cf.~Subsection~\ref{subsec:deduc_thm}). This theorem involves that each judgment in a system of terms must be consistent and logically independent, i.e.,~the judgment must be based on a calculus system, to conclude that the derived judgment is deducible (cf.~\citet[Chap.~11.]{ebb:07}). In contrast to these premises we observe that the negation of $A \Rightarrow B$, the expression $\neg(A \Rightarrow B)$ cannot be deducible. Analogously, we know that $A \Rightarrow A$ is a deduction, but then the negation of $A \Rightarrow A$ i.e., $\neg(A \Rightarrow A)$, is not deducible as well. Thus, the author claims that one gets something consistent that is based on inconsistencies, which is absurd. In fact, this says that from an inconsistent theory one can deduce everything. The author is captured by the {\bfseries Principle of Explosion} -- ex falso sequitur quodlibet,~i.e., from falsehood anything follows (cf.~Section~\ref{sec:logstat}) -- that destroys the entire structure of his game model. This is due that the premise $\neg(A \Rightarrow B)$ is inconsistent, it cannot be true, then drawing from it a conclusion is meaningless, and anything can be concluded. Hence, from an inconsistency any statement can be proven making the concepts of truth and falsehood irrelevant with the consequence that the underlying game model must be useless. Finally, the author has failed to establish that the candidate of a calculus system is logically independent, which means in that case, he failed to show that the prerequisites $A$ and $B$ are logically independent. None of the applied arguments of the author allow us to conclude that $(A \land \neg B) \Rightarrow (A \land \neg A)$ is provable, and therefore a valid logical sentence.     

We now outline why his purported proof of the {\itshape ``sufficiency case''} of Proposition 3.1 must be wrong. Similar as in  other examples from the literature (cf.~\citet{mei:15,mei:16,mei:16b,mei:17,mei:18a,mei:18b}), he uses elements from a material implication for establishing the logical equivalent proposition {\itshape if $A$ then $B$}. This author starts with $A \land \neg B$ to perform this kind of proof to get a contradiction in order to conclude that the implication $A \Rightarrow B $ is drawn. Therefore, the author does not recognize that whenever a valid premise $A \land \neg B$ implies something false like $\neg A$, one cannot get a valid logical sentence. As we stressed out in the previous paragraph, he deduces from a non-deducible sentence something that is non-deducible, which cannot be consistent. In this case, the implication must be a falsehood, which says that the clause is not provable. From which immediately follows -- and which is the sole conclusion that can be drawn from this non provable clause -- that a Nash equilibrium of the normal form oligopoly game $\Gamma^{\mathcal{P}} := (\mathcal{P}, (X^{S},\pi_{S})_{S \in \mathcal{P}})$ does not always exist.    

When we are going into the details we recognize that the author applies the prerequisite $A$ of the positive statement and $\neg B$ in order to prove the contrapositive statement {\itshape if $\neg B \Rightarrow \neg A$}. For doing so, he assumes that the payoff vector $\hat{x}^{\mathcal{P}} \in X^{\mathcal{P}}$ is a Nash equilibrium of the normal form oligopoly game $\Gamma^{\mathcal{P}} = (\mathcal{P}, (X^{S},\pi_{S})_{S \in \mathcal{P}})$, that is, premise $A$ holds. It should be clear that assuming the existence of Nash equilibrium for game $\Gamma^{\mathcal{P}}$ is not a very elegant proof technique to ignore the aggregation issue across firms, which, nevertheless, requires more sophisticated methods as those used by the author to overcome its problematic nature. In the next step, it is then assumed that the strategy profile $\hat{x} = (\hat{x}_{S})_{S \in \mathcal{P}} \in X_{N}$ is not a Nash equilibrium of the normal form oligopoly game $\Gamma = (N, (X_{i},\pi_{i})_{i \in N})$ under $\mathcal{P}$, i.e., premise $B$ is false. Premise $A$ is then used in his proof to construct in a first step the vector $\hat{x}$, and finally to construct the contradiction that $\hat{x}^{\mathcal{P}} \in X^{\mathcal{P}}$ is not a Nash equilibrium ($\neg A$). In effect, he has shown that $A \land \neg B \Rightarrow A \land \neg A$ is not consistent. As a consequence, the clause $A \Rightarrow B$ must be false too, in accordance with $(A \land \neg B \Rightarrow A \land \neg A) \equiv (A \Rightarrow B)$. Hence, $(A \land \neg B \Rightarrow A \land \neg A)$ is not consistent, and therefore not provable. This is the only possible conclusion w.r.t.~the imposed regular conditions from industrial organization, that do not induce a well-behaved aggregate profit function on the cartel level. Ignoring the well-known issue of inheriting individual properties to an aggregated level, the author incorrectly applied $(\phi \Rightarrow \bot) \Leftrightarrow \neg \phi$ while referring to fatal arguments.      
 
For completeness, we just want to mention that the same misguided line of argument is also given for the {\itshape ``necessity case''}. There, he is not aware about the following logical equivalence $(\neg A \land B \Rightarrow \neg A \land A) \equiv (B \Rightarrow A)$. No wonder that he shows that the truth $\neg A \land B$ implies a falsehood $\neg A \land A$, which is as well a wrong logical clause. It follows that $B \Rightarrow A$ is not provable. 
 
In summary, he has shown in both cases the exact opposite of what he had claimed to prove while not following a good practice of logical reasoning and fading out crucial aspects from the aggregation issue across firms. As a consequence,~\citeauthor{lard:12} has disproved his own Proposition 3.1.     
 
In the sequel, we show what will happen if we apply a proof by contraposition $\neg B \Rightarrow \neg A$ for the {\itshape ``sufficiency case''} in order to see where we run into problems. But then the starting point of the proof has to be the assumption that the payoff vector $\hat{x} \in X_{N}$ is not a Nash equilibrium of the normal form oligopoly game $\Gamma = (N, (X_{i},\pi_{i})_{i \in N})$ under $\mathcal{P}$ ($\neg B$), which implies by imposing the correct assumption like quasi-concavity on the profit function $\pi_{i}$ in order to guarantee existence of an equilibrium that 
 \begin{equation*}
  \sum_{i \in S}\, \pi_{i}(\hat{x}_{S},\hat{x}_{-S}) \le   \sum_{i \in S}\, \pi_{i}(\check{x}_{S},\hat{x}_{-S}),
\end{equation*}
is true. In this case, Formula (11) of~\citet[p.~394]{lard:12} -- that gives the definition of the aggregate cost function $C_{S}:X^{S} \rightarrow \mathbb{R}_{+}$ -- does not ensure continuity of $C_{S}$ implying for the payoff vector $\hat{x} \in X_{N}$ that 
\begin{equation*}
  \sum_{i \in S}\, C_{i}(\hat{x}_{i}) = C_{S}(\hat{x}^{S}).
\end{equation*}
cannot be anymore estimated. This is due that it cannot be supposed -- without committing severe logical failures due to the violation of imposing necessary and sufficient properties on firms profit functions in order to get a representation of an aggregate profit function that inherits their properties -- that $\hat{x}^{\mathcal{P}} \in X^{\mathcal{P}}$ is a Nash equilibrium. We have to stress the fact that the application of the Deduction Theorem (cf.~Section~\ref{subsec:deduc_thm}) requires in that case the existence (provability) of a Nash equilibrium of the normal form game $(\mathcal{P}, (X^{S},\pi_{S})_{S \in \mathcal{P}})$, which means that imposing its existence just by an assumption is not a permissible conduct. We have seen by the foregoing consideration that such a Nash equilibrium cannot be assured by the regular conditions imposed by the game model. Hence, imposing its existence must lead to a wrong conclusion. Existence of an equilibrium cannot be assured, as a consequence, it is also not anymore clear that 
\begin{equation*}
   \pi_{S}(\hat{x}^{\mathcal{P}}) < \pi_{S}(\check{x}^{S}, \hat{x}^{-S}),
\end{equation*}
is satisfied as it was claimed by~\citet[p.~395]{lard:12}. This inequality can only be obtained when the author can establish by some logical inference in consideration of the aggregation issue that $\hat{x}^{\mathcal{P}} \in X^{\mathcal{P}}$ is a Nash equilibrium of the normal form oligopoly game $\Gamma^{\mathcal{P}} = (\mathcal{P}, (X^{S},\pi_{S})_{S \in \mathcal{P}})$ ($A$ is valid), but not by an assumption. Moreover, Corollary 3.2~is not correct either, implying in connection with the disproof of Proposition~3.1 that the TU game in $\gamma$-characteristic function form is not well-defined. Again, the results of the article are devalued according to these logical flaws. 

\subsection{Not well-behaved Aggregate Behavior across Firms}
\label{sub2:exp_nwagb}

To see that a TU-game in $\gamma$-characteristic function form without transferable technologies might not be well-defined, we introduce a Cournot oligopoly situation, denoted $\langle N, (\omega_{k})_{k \in N},(c_k)_{k \in N}, p \rangle$, with five firms $N=\{1,2,3,4,5\}$ producing a homogeneous good and having production capacities given by the vector of capacities $\{\omega_{1},\omega_{2},\omega_{3},\omega_{4},\omega_{5}\}=\{2.6, 0.1, 2, 15, 20\}$. The individual marginal costs of the firms are given by the cost vector $\{c_{1},c_{2},c_{3},c_{4},c_{5}\}=\{1/8, 5/2, 5, 1/24, 1\}$. Thus, the cost functions are increasing in its arguments. Furthermore, we assume that the parameters of the concave inverse demand function $p(X)=a - b \cdot X^{2}$ are given by $\{a,b\}=\{120,1\}$ with joint production of $X=\sum_{i \in N}\,x_{i}$. Then the profit function is specified for each firm $i$ by $\pi_{i}(x):= p(X) \cdot\, x_{i} - c_{i}\cdot\,x_{i}$. Notice, that the inverse demand function $p(X)$ is strictly decreasing and concave in its arguments within the relevant domain. 
 
\begin{figure}[H]
\ifpdf
    \includegraphics[height=9.0cm, width=14.5cm]{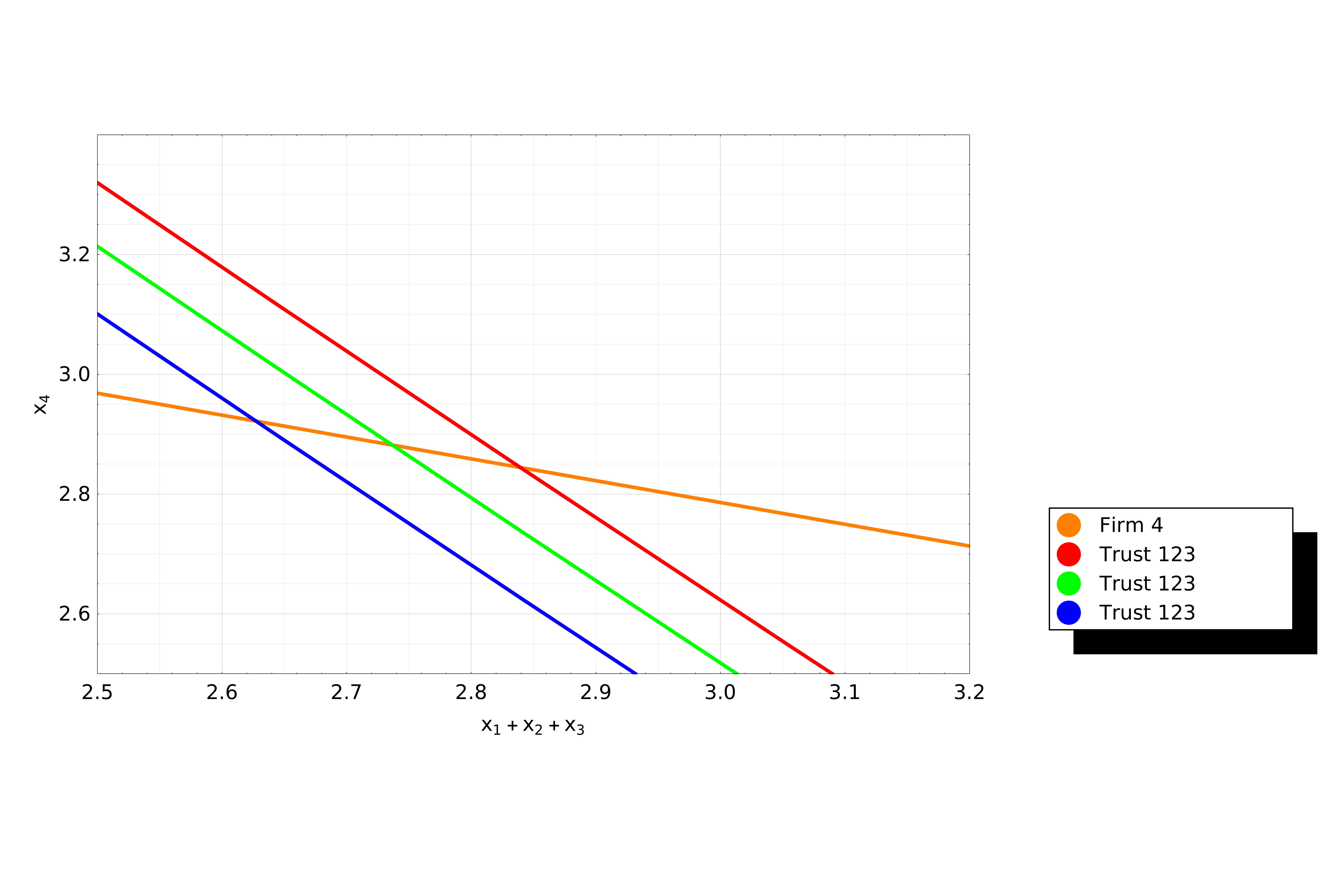} 
\else
    \includegraphics[height=9.0cm, width=14.5cm]{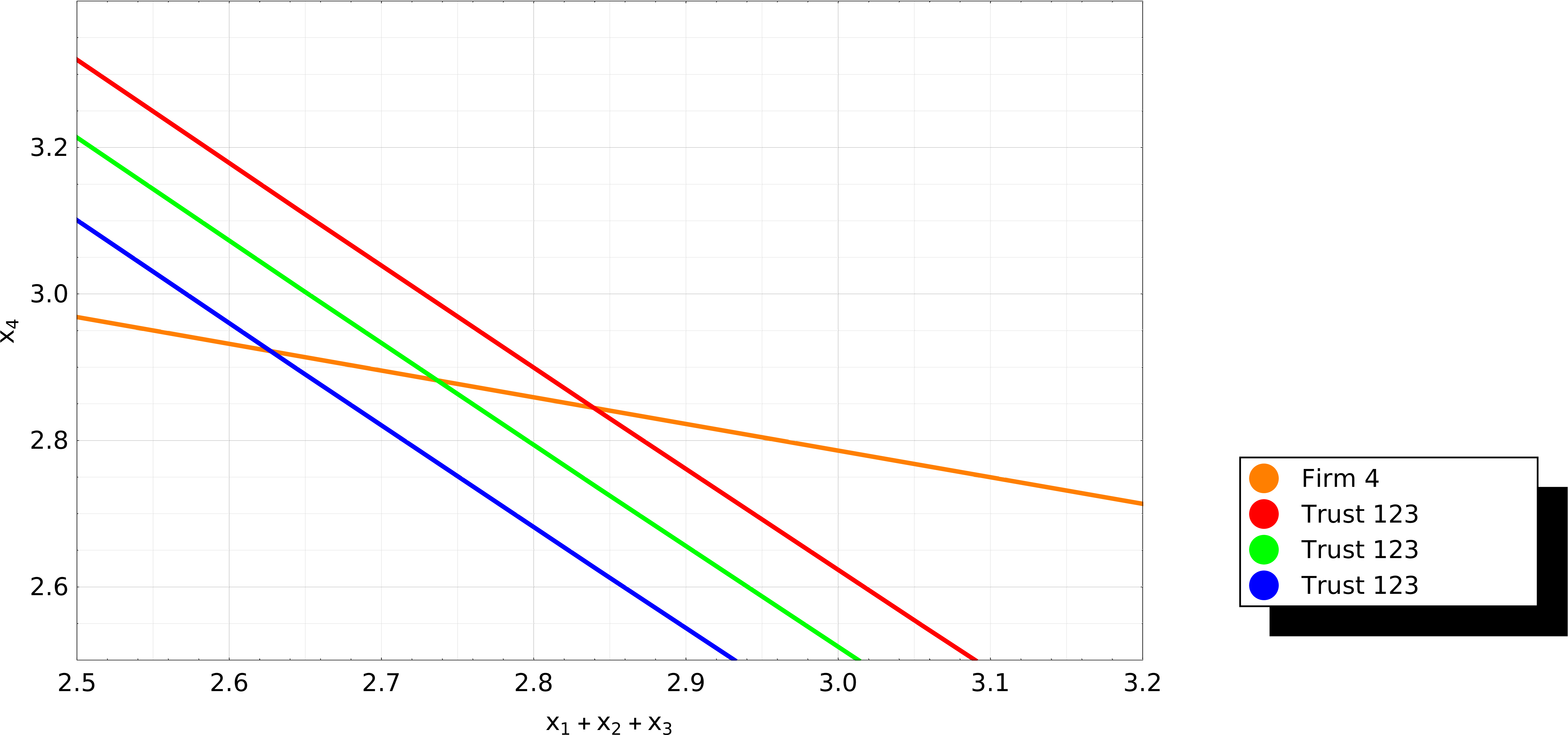} 
\fi
    \caption{Individual Best Reply Functions of Trust $\{1,2,3\}$ and Firm $4$} 
    \label{fig:brf1}
\end{figure}

\begin{figure}[H]
\ifpdf
    \includegraphics[height=9.0cm, width=14.5cm]{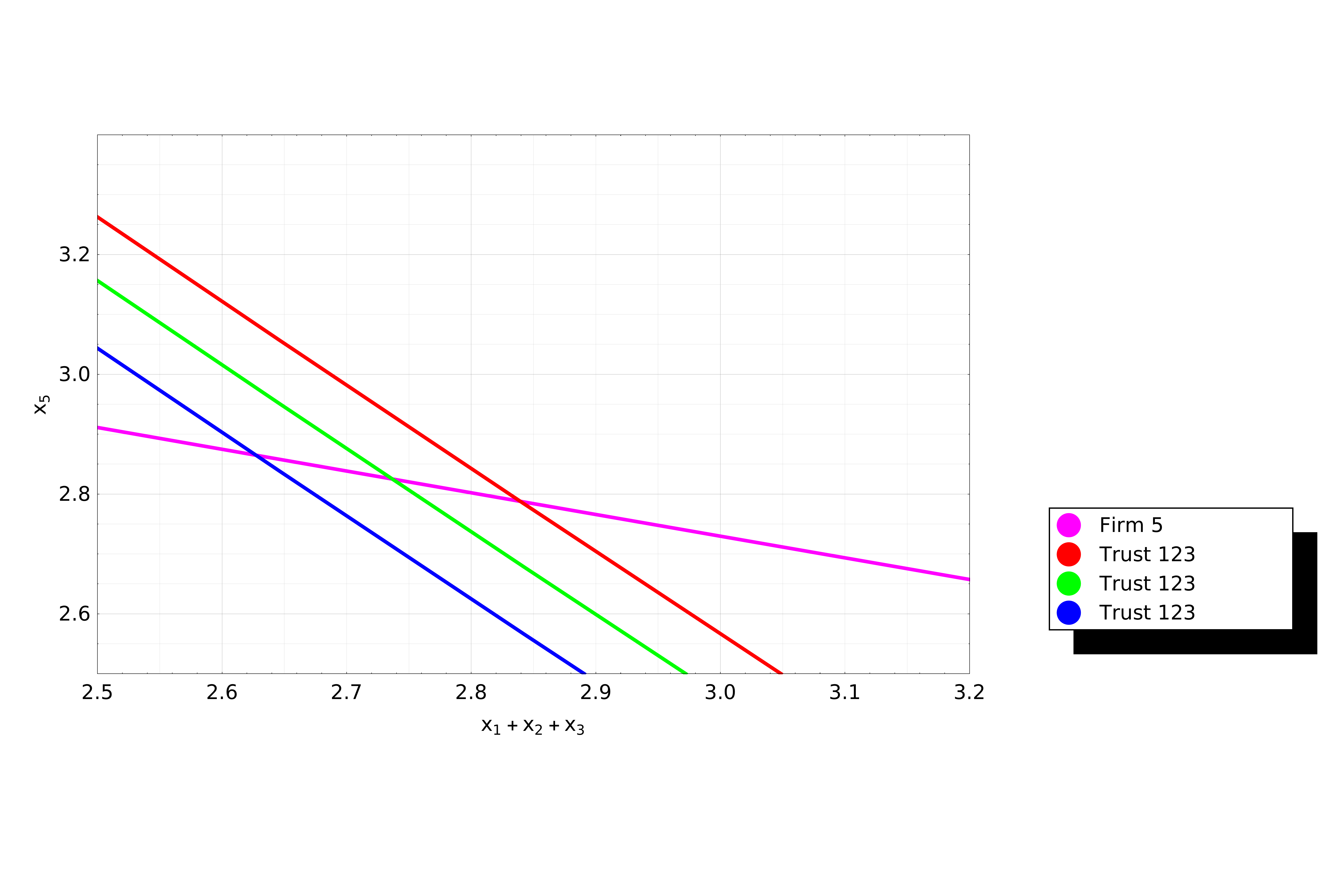} 
\else
    \includegraphics[height=9.0cm, width=14.5cm]{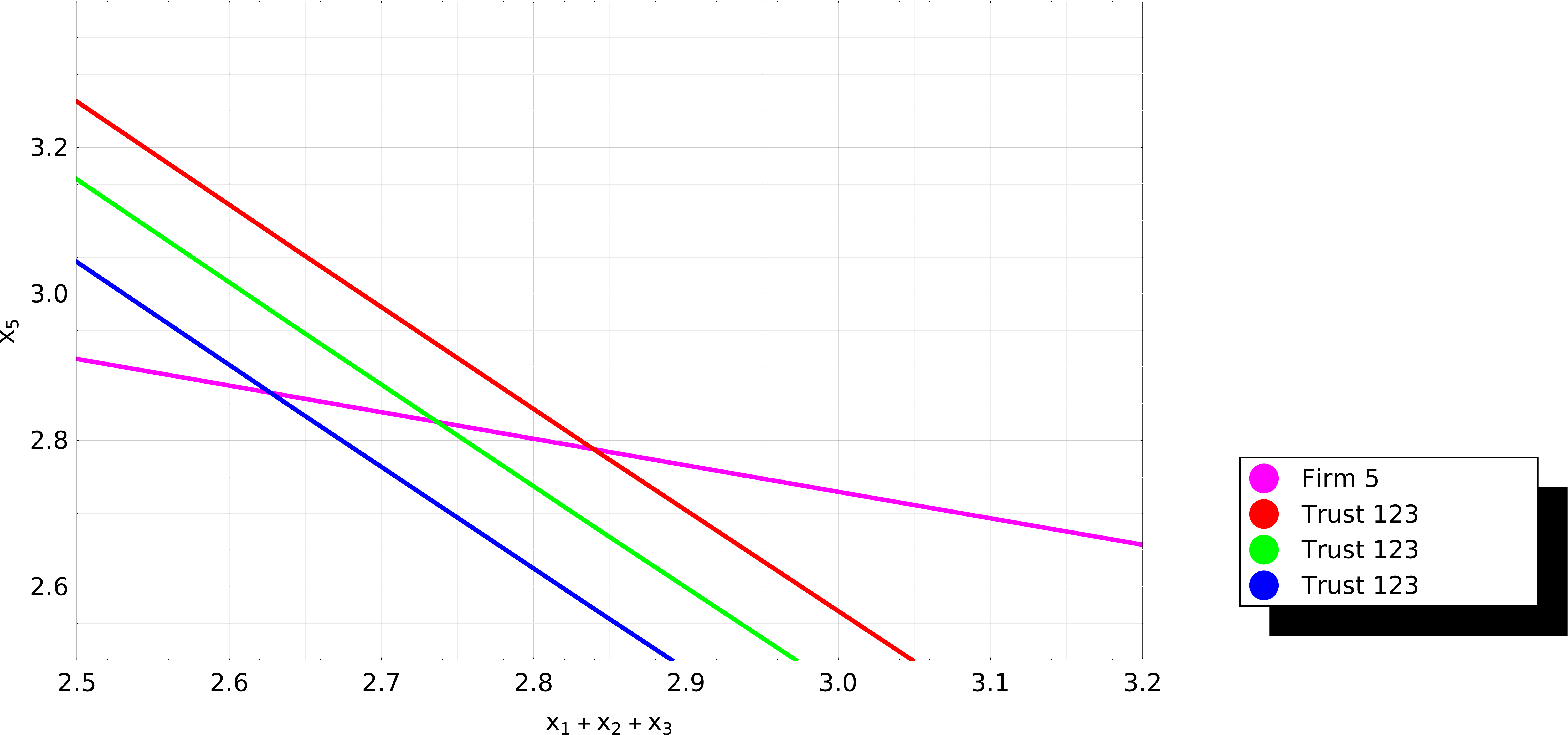} 
\fi
    \caption{Individual Best Reply Functions of Trust $\{1,2,3\}$ and Firm $5$} 
    \label{fig:brf1b}
\end{figure}

\begin{figure}[H]
\ifpdf 
   \includegraphics[height=9.0cm, width=14.5cm]{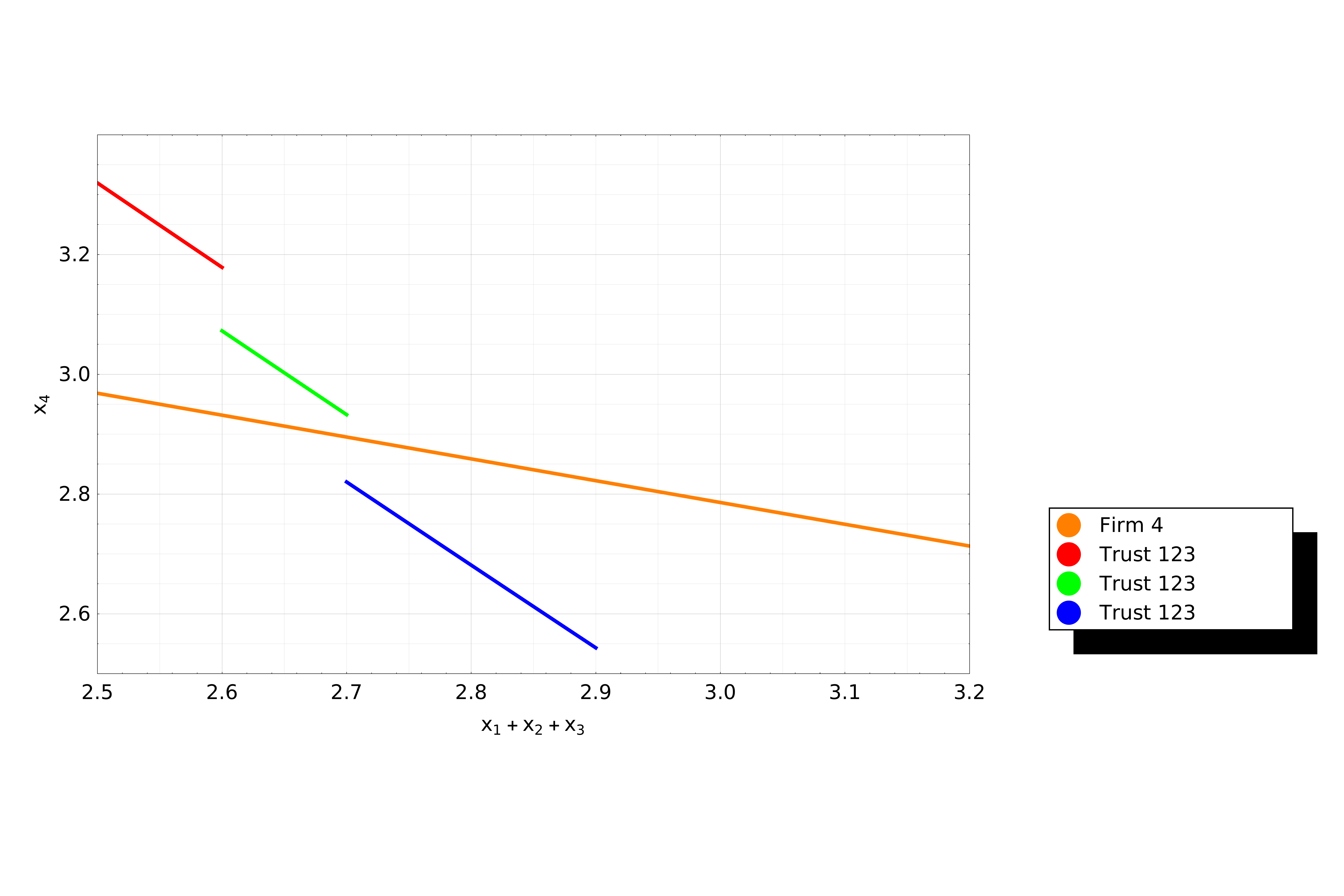} 
\else
   \includegraphics[height=9.0cm, width=14.5cm]{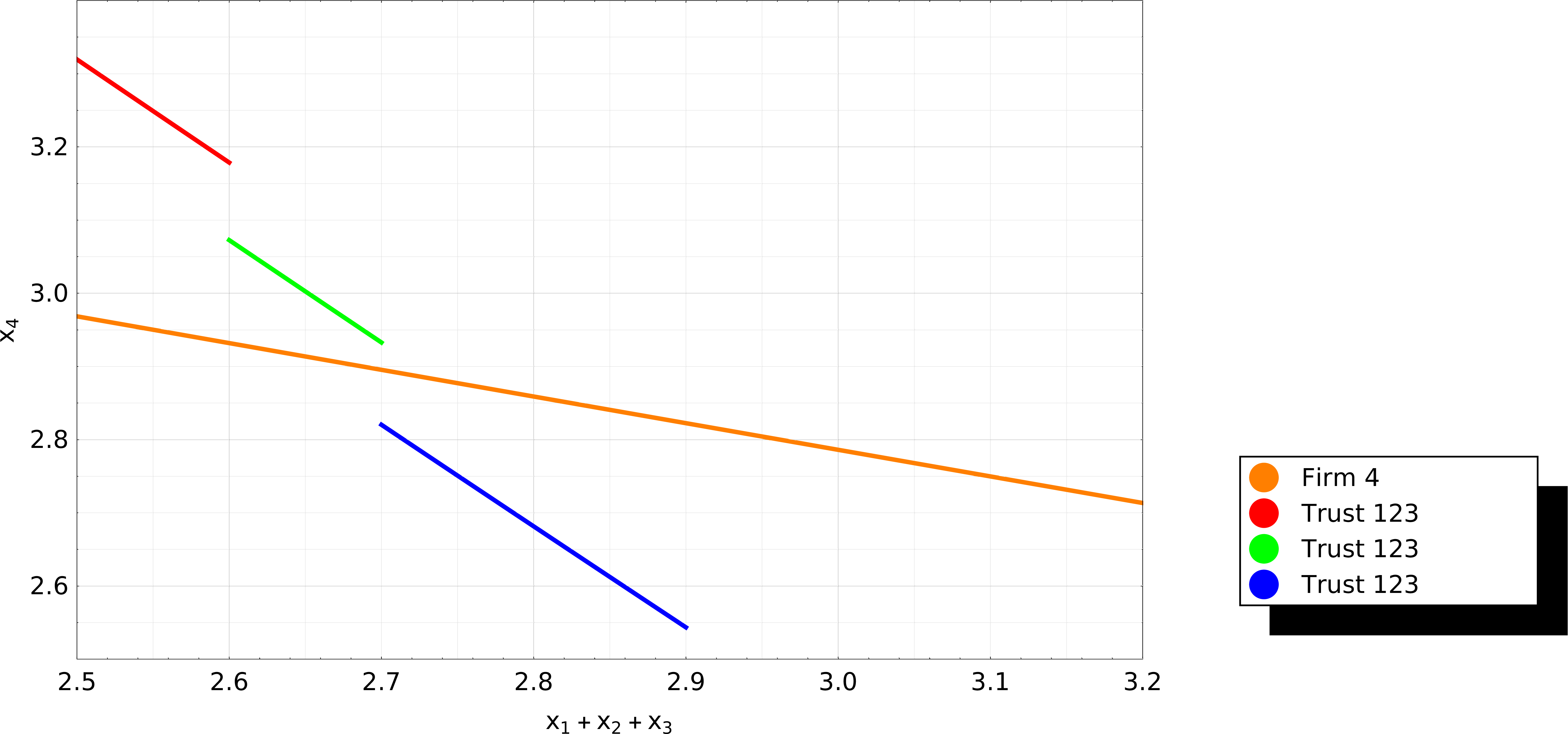} 
\fi
    \caption{Best Reply Functions of Trust $\{1,2,3\}$ and Outsider $4$} 
    \label{fig:brf2}
\end{figure}

\begin{figure}[H]
\ifpdf 
   \includegraphics[height=9.0cm, width=14.5cm]{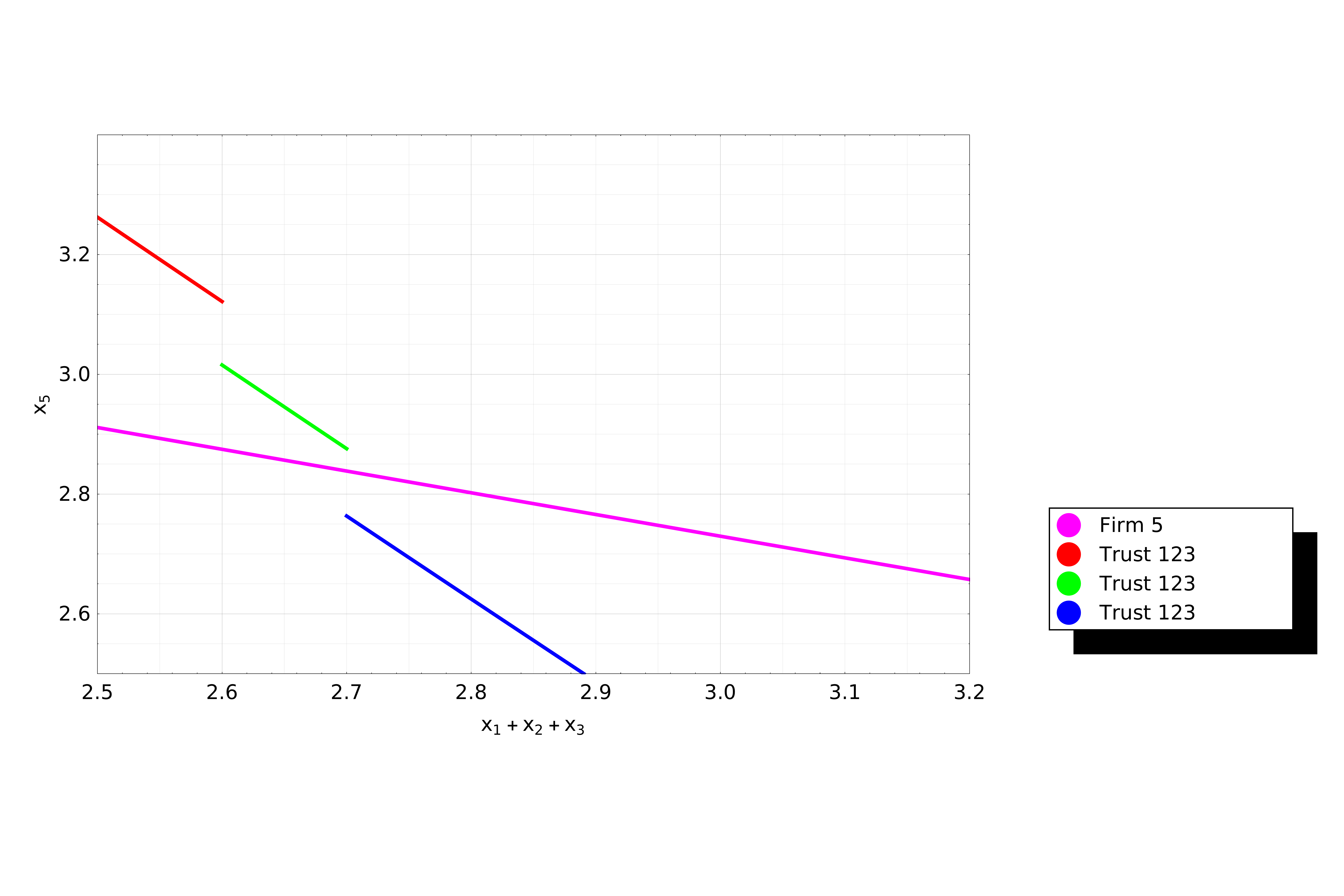} 
\else
   \includegraphics[height=9.0cm, width=14.5cm]{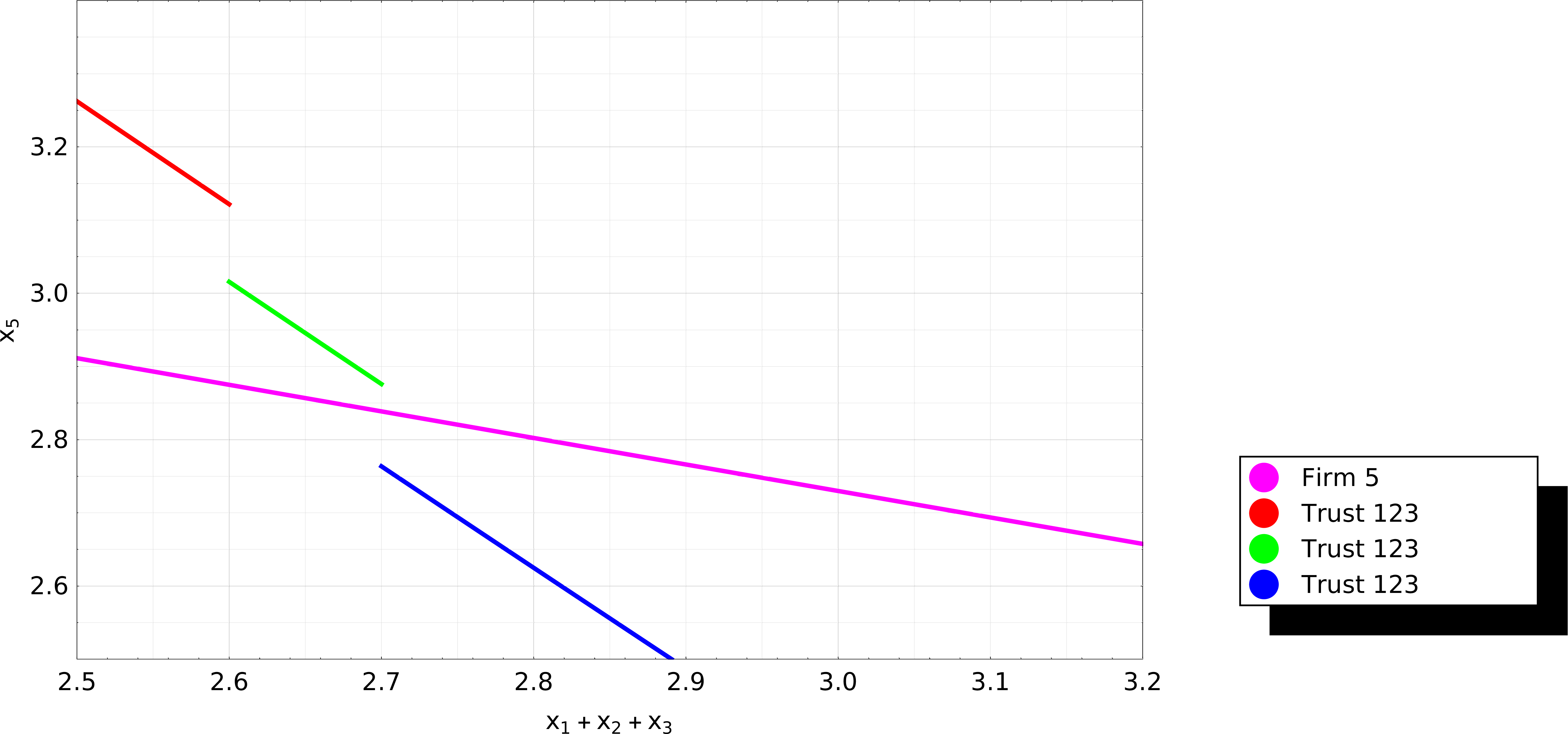} 
\fi
    \caption{Best Reply Functions of Trust $\{1,2,3\}$ and Outsider $5$} 
    \label{fig:brf2b}
\end{figure}

To observe that no Nash-equilibrium can be guaranteed for the game $(\mathcal{P}_{\gamma}, (X^{S},\pi_{S})_{S \in \mathcal{P}_{\gamma}})$, we consider the coalition structure $\mathcal{P}_{\gamma} = \{\{1,2,3\},\{4\},\{5\}\}$. The marginal profit function of coalition/trust $\{1,2,3\}$ is defined by its f.o.c., and it is given by  
\begin{equation*}
  \Delta\,\pi_{\{1,2,3\}}(\bar{X}) = 
  \begin{cases}
    \frac{1}{16} \left(\frac{5773}{6} -10 \bar{X}^2-10 \sqrt{\bar{X}^{2}+\frac{5735}{6}} \bar{X}\right)  & \text{if}\,\bar{X} \in [0,2.6] \\
    \frac{1}{16} \left(\frac{5545}{6} -10 \bar{X}^2-10 \sqrt{\bar{X}^{2}+\frac{5735}{6}} \bar{X}\right) & \text{if}\,\bar{X} \in (2.6,2.7] \\
    \frac{1}{16} \left(\frac{5305}{6}-10 \bar{X}^2-10 \sqrt{\bar{X}^{2}+\frac{5735}{6}} \bar{X}\right)& \text{if}\,\bar{X} \in (2.7,4.7],
  \end{cases}
\end{equation*}
with $\bar{X}=\sum_{i \in \{1,2,3\}}\,x_i$. Similar, the individual best replies of outsider firms $4$ and $5$ are quantified through
\begin{equation*}
  \begin{split}
  B_{4}(\bar{X}) & = \frac{3}{5735} \left(\frac{1927}{8} \sqrt{\bar{X}^{2}+955.833}-\frac{89849}{125} \bar{X}\right) \quad\text{if}\,\bar{X} \in [0,5] \\
  B_{5}(\bar{X}) & = \frac{3}{5735} \left(\frac{28208}{119} \sqrt{\bar{X}^{2}+955.833}-\frac{85080}{119} \bar{X}\right) \quad\text{if}\,\bar{X} \in [0,5].
  \end{split}
\end{equation*}
Taken the best response function of outsider $5$ as given, the inverse best response of the trust $\{1,2,3\}$ w.r.t.~the output level $x_{4}$ of outsider $4$ can be determined from the f.o.c. Analogous arguments let us determine the inverse best response of the trust $\{1,2,3\}$ w.r.t.~the output level $x_{5}$ of outsider $5$. We have depicted the individual best reply functions in Figures~\ref{fig:brf1} and~\ref{fig:brf1b}, whereas Figures~\ref{fig:brf2} and~\ref{fig:brf2b} show the aggregate best response of cartel $\{1,2,3\}$, and the individual ones of outsider firms $4$ and $5$. 

That we cannot guarantee a Nash-equilibrium in the game model is due that the system of equations derived by the first-order condition is not independent, and therefore underdetermined, i.e., the system has fewer independent equations than variables. Such a system has no solution. This can be observed by the above system of best reply functions as well by Figures~\ref{fig:brf1} and~\ref{fig:brf1b}. The individual best replies of firms of trust $\{1,2,3\}$ are shifted by constants downward, they are not independent. To guarantee in this case an equilibrium it is enough to assume that the most efficient firm $1$ of trust $\{1,2,3\}$ has a sufficient large capacity. Thus, we can assure an intersection of the best reply function of coalition $\{1,2,3\}$ with that of outsider firms $4$ and $5$ (cf.~Figures~\ref{fig:brf1} and~\ref{fig:brf1b}). But then there is no need to focus on oligopoly TU-games without transferable technologies. Hence, to assure an equilibrium, we have either to consider a game model with transferable technologies, i.e., all firms of a trust can rely on the most efficient production technology, or we have to introduce a substitution rate among the goods violating the homogeneous good assumption. While incorporating a substitution rate under such a game model without synergy effects among the firms, we can guarantee a Nash-equilibrium, but this would change the theoretical framework in which firms are involved, and the $\gamma$-core might be empty. Moreover, trying to overcome the issue of a non-existing equilibrium while incorporating capacity constraints, and making the strategy set compact is not enough to assure an equilibrium as we can observe by Figures~\ref{fig:brf2} and~\ref{fig:brf2b} 

Thus, if we choose the above capacity levels, then the best reply of trust $\{1,2,3\}$ becomes discontinuous on the relevant domain with the consequence that we do not observe anymore an intersection of best reply functions (cf.~Figures~\ref{fig:brf2} and~\ref{fig:brf2b}).\footnote{Note that the best response function of trust $\{1,2,3\}$ cannot connect the three downward-sloping lines by vertical lines indicating that it is then an upper hemi-continuous best reply correspondence rather than a best reply function. The imposed regular conditions in the literature (cf.~\citet{zhao:16f,mei:17f}) of getting a $\gamma$-characteristic function induce only a best response function and not an u.h.c.~best response correspondence for the trust. On the trust's production level $2.6$ and $2.7$ we observe points of discontinuity for the aggregate cost function implying that the aggregate profit function cannot be continuous either on these levels. Hence, there can only be a jump happen when the production cross over from firm $1$ to $2$ and from firm $2$ to $3$ indicating an abrupt transition of best response. Thus, trust $\{1,2,3\}$ does not have a constant best response within the interval of $[3.07,3.17]$ or $[2.82,2.93]$ w.r.t.~$x_{4}$. The same argument holds for the inverall $[3.02,3.12]$ or $[2.76,2.87]$ w.r.t.~$x_{5}$.} There is no intersection of best replies, therefore no Nash-equilibrium can exist. Hence, no $\gamma$-characteristic value of coalition $\{1,2,3\}$ can be determined, the associated $\gamma$-characteristic function is void. This establishes that the game model is not correctly specified, because of the mentioned aggregation issue across firms. By this counter-example we have demonstrated that the purported proof of~\citet[Proposition 3.1]{lard:12} is logical flawed and is therefore false.        
 
\section{Concluding Remarks}
\label{sec:rem}
We presented a critical review of the common practice in game theory to run a purported proof by contradiction while relying on the Deduction Theorem. We have argued that the arguments which have been used in its favor are violating good practice of logical reasoning, and are therefore wrong. To underpin our consideration, we discussed an example from the literature that obviously conflicts with the well-known aggregation issue. We express our hope that the presented arguments are a small contribution toward a higher reliability and consistency of the published results.    
 
\pagebreak
\pagebreak

\pagebreak
\pagebreak

\pagestyle{scrheadings} \chead{\empty}  
\footnotesize

\bibliography{deduction}

\end{document}